\documentclass[letterpaper, 10 pt, conference]{ieeeconf}  

\IEEEoverridecommandlockouts                              

\title{\LARGE \bf
Graph-Based Heuristics for Minimum Constraint Removal in Cluttered Robot Motion Planning
}

\author{Antony Thomas$^{1}$, Fulvio Mastrogiovanni$^{2}$, and Marco Baglietto$^{2}$
\thanks{$^{1}$Robotics Research Center, IIIT Hyderabad, 500032, Hyderabad, India.
{Email: antony.thomas@iiit.ac.in}}%
\thanks{$^{2}$Department of Informatics, Bioengineering, Robotics, and Systems Engineering, University of Genoa, 16145, Genoa, Italy. 
{Emails: fulvio.mastrogiovanni@unige.it, marco.baglietto@unige.it}.}%
}

\usepackage{url}
\usepackage{graphicx}        
\usepackage{multicol}        

\usepackage{amsmath,bm,amsfonts,amssymb}

\usepackage{commath}      
\usepackage{subfig}
\usepackage[font=small,labelfont=bf]{caption}

\usepackage{floatrow}
\usepackage{tabularx}
\usepackage{float}
\usepackage{upgreek}
\usepackage[euler]{textgreek}
\usepackage{wrapfig,booktabs}

\usepackage{algorithm}
\usepackage[noend]{algpseudocode}
\usepackage{colortbl}  
\usepackage{multirow}
\usepackage{listings}
\usepackage{lipsum}
\usepackage{textcomp} 

\usepackage[nocomma]{optidef}  
\usepackage{tikz} 
\usepackage{tikzscale}

\newtheorem{definition}{Definition}

\makeatletter
\newcommand{\clearsubcaptcounter}{\setcounter{sub\@captype}{0}}
\makeatother

\newcolumntype{C}[1]{>{\centering\arraybackslash}p{#1}}

\definecolor{revisioncolor}{rgb}{0.1,0.1,1}

\algnewcommand\algorithmicforeach{\textbf{for each}}
\algdef{S}[FOR]{ForEach}[1]{\algorithmicforeach\ #1\ \algorithmicdo}

\algnewcommand\algorithmicifempty{\textbf{if empty}}
\algdef{S}[FOR]{IfEmpty}[1]{\algorithmicifempty\ #1\ \algorithmicdo}
\usetikzlibrary{positioning}

\begin{document}

\maketitle
\thispagestyle{empty}
\pagestyle{empty}

\begin{abstract}
Humanoid and service robots operating in cluttered human environments may encounter motion planning problems that are infeasible when all surrounding objects are treated as fixed obstacles. 
The Minimum Constraint Removal problem, which we refer to as \textsf{MCR}, addresses this situation by identifying the minimum number of constraints, such as movable objects, whose removal enables a collision-free path. 
However, computing an optimal \textsf{MCR} solution is NP-hard. 
In this work, we propose a graph-based heuristic pipeline that combines bidirectional feasible-set generation with fixed-length searches over reduced \textsf{MCR} graphs. 
We further derive an extended feasible-set generator for instances in which the
candidates produced by the first generator do not contain a complete optimal removal set.
Across six benchmarks, the evaluated pipeline matches the exact solution in three cases and, in the remaining cases, selects at most $31.25\%$ more obstacles than the optimum, while being between $3.4$ and $33.7$ times faster than exact search.
In one of the benchmarks, involving a manipulation task amidst clutter, selecting two rather than seven obstructing objects reduces the combined \textsf{MCR}, inverse-kinematics, and motion planning time from $84.72$~s to $30.59$~s. 
These results indicate that the proposed approach provides a practical compromise between computational efficiency and solution quality for motion planning among movable obstacles.
\end{abstract}


\section{Introduction}
\label{sec:intro}
Humanoid and service robots are expected to operate in human environments that are not arranged specifically for them. 
In such environments, a motion planning problem may be infeasible when all surrounding objects are treated as \textit{fixed} obstacles, even though some of them could be actually \textit{moved}. 
For example, a robot reaching into a shelf may need to relocate objects that obstruct its arm, while a robot navigating through a domestic environment may need to reposition a chair or open a door. 
Because each additional interaction requires perception, manipulation planning, and execution, it is desirable to identify the smallest set of objects whose relocation restores motion feasibility.

The Minimum Constraint Removal problem, henceforth referred to as \textsf{MCR}, introduced in robotics by Hauser~\cite{hauser2014IJRR}, formalizes this objective. 
Given a start and a goal configuration, \textsf{MCR} seeks a path that intersects the minimum number of distinct constraints in configuration space. 
When these constraints represent movable objects, the solution identifies a minimum-cardinality set of objects whose removal would make the corresponding motion feasible. 
In this sense, \textsf{MCR} provides a compact explanation of motion planning infeasibility and identifies which elements of the environment must be modified before a feasible motion can be found.

This capability is relevant to several planning problems. 
In task and motion planning~\cite{kaelbling2013IJRR, srivastava2014ICRA, dantam2016RSS, garrett2018IJRR, thomas2021RAS}, a task-level plan may need to be revised when its geometric realization is infeasible. 
Minimum Constraint Displacement methods~\cite{hauser2013RSS, thomas2022IAS, thomas2023ICRA} instead determine how far obstacles should be displaced to recover feasibility. 
Related challenges arise in Navigation Among Movable Obstacles~\cite{stilman2005IJHR, nieuwenhuisen2008WAFR, van2009WAFR} and in manipulation and rearrangement planning~\cite{stilman2007ICRA, dogar2011RSS, krontiris2015RSS, karami2021AIIA}, where objects are moved or rearranged to create a path to a previously unreachable goal. 
In contrast to methods that explicitly plan \textit{how} and \textit{where} each object should be relocated, the present work focuses on the preceding selection problem, that is, \textit{determining which obstacles need to be removed}.

The \textsf{MCR} problem is NP-hard by reduction from \textsc{Set Cover}~\cite{hauser2014IJRR}.
The problem remains NP-hard for convex polygonal obstacles~\cite{erickson2013AAAI} and for line-segment obstacles~\cite{alt2011EuroCG,yang2012PHD}.
Consequently, several studies have investigated approximate solutions~\cite{gorbenko2015AIP, krontiris2017AR, eiben2018AAAI, kumar2019AEA, bandyapadhyay2020CG}.
For restricted geometric settings, Bandyapadhyay \textit{et al.}~\cite{bandyapadhyay2020CG} establish an $O(\sqrt{n})$ approximation for rectilinear polygons and disks. 
When \textsf{MCR} is represented as a minimum-color path problem on an edge-colored graph, an $O(n^{2/3})$ polynomial-time approximation is available~\cite{kumar2019AEA}, with $n$ denoting the number of graph vertices. 
Nevertheless, practical robot instances still present a difficult trade-off, that is, exact graph search can be computationally expensive, whereas aggressive greedy pruning can return solutions containing substantially more obstacles. 
This difficulty stems from the non-additive nature of the objective, since encountering the same obstacle multiple times along a path incurs a cost only once.

In this paper, we introduce a graph-based heuristic framework for efficiently computing feasible, low-cardinality solutions to \textsf{MCR}, implemented as two feasible-set generators. 
The first feasible-set generator, referred to as \textsf{GFS-1}, performs bidirectional exploration from the start and goal vertices and uses obstacle labels shared by the two search frontiers to construct candidate removal sets. 
These candidates are subsequently refined by searching for fixed-length paths on reduced \textsf{MCR} graphs, via a \textsf{Refine} procedure.
The second feasible-set generator, called \textsf{GFS-2}, extends the search to instances in which the candidates produced by the first generator may not contain a complete optimal removal set.
The main contributions of this work are:
\begin{itemize}
\item
the bidirectional feasible-set generator \textsf{GFS-1}, and a reduced-graph refinement procedure for computing feasible, low-cardinality obstacle-removal sets;
\item
the extended generator \textsf{GFS-2}, which addresses candidate-coverage cases not handled by \textsf{GFS-1}, together with an analysis of its computational complexity;
\item an evaluation of $\textsf{GFS-1}+\textsf{Refine}$ on six benchmarks, including a manipulation scenario, in which it matches the exact solution in three cases, selects at most $31.25\%$ more obstacles than optimal in the remaining cases, and achieve speed-up factors between $3.4$ and $33.7$ over exact search.
\end{itemize}

The manipulation benchmark further shows that reducing the number of selected obstacles can substantially decrease downstream planning effort.
Compared with greedy search, our approach reduces the number of selected objects from seven to two and the combined \textsf{MCR}, inverse-kinematics, and motion-planning time from $84.72$~s to $30.59$~s.

\section{Preliminaries}
\label{sec:prelim}
\begin{figure}[t!]
\centering
\includegraphics[trim=92 110 70 6,clip,scale=0.95]{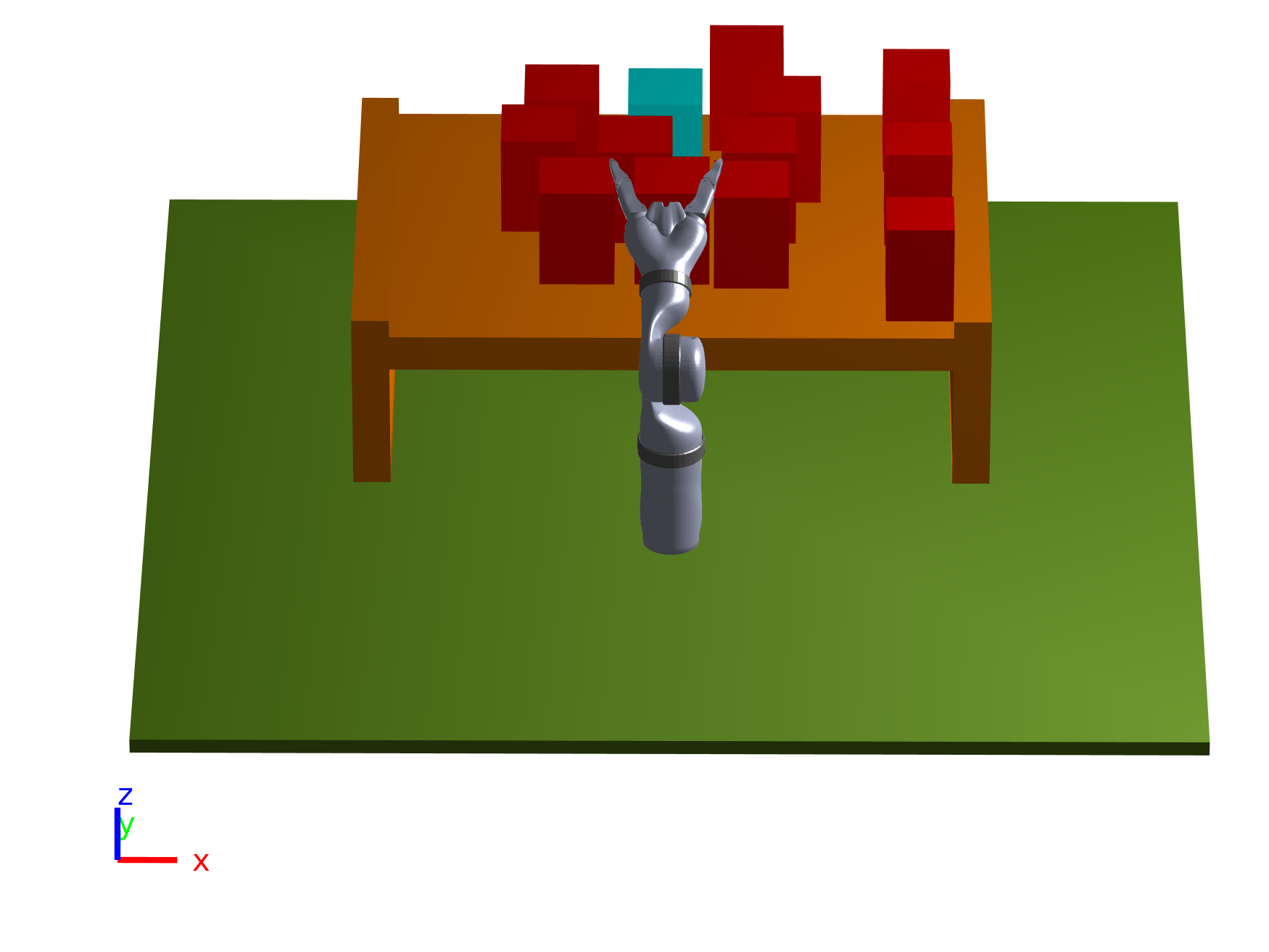}
\caption{
An \textsf{MCR} instance for a manipulation task. 
The robot must reach and grasp the cyan object, but the red objects make the corresponding motion planning problem infeasible.}
\label{fig:prelim0}
\end{figure}

Figure~\ref{fig:prelim0} illustrates a manipulation problem in which a robot must reach and grasp a target object in a cluttered workspace.
No collision-free path exists when the red objects are treated as fixed obstacles.
However, assuming that these objects are movable, the objective is to identify a minimum-cardinality subset of red objects whose relocation would restore motion feasibility.
In this work, \emph{removing} an obstacle is an abstraction for relocating the
corresponding object \textit{outside} the region relevant to the planned motion.
The relocation motion itself is not computed by \textsf{MCR}.

\begin{definition}
Let
$\mathcal{C}\subseteq\mathbb{R}^{d}$
be a configuration space, let
$\mathcal{O} = \{o_1, \ldots, o_n\}$ be a set of configuration space obstacle regions, and let
$s,g\in\mathcal{C}$
be the start and goal configurations.
For a candidate removal set $S\subseteq\mathcal O$, removing $S$ leaves the obstacle set $\mathcal O\setminus S$.
The Minimum Constraint Removal problem is
\begin{equation}
\begin{aligned}
S^\star \in \arg\min_{S\subseteq\mathcal{O}}\;& |S| \\
\mathrm{s.t.}\quad
&\exists\,\tau:[0,1]\rightarrow
\mathcal{C}\setminus
\bigcup_{o\in\mathcal{O}\setminus S}o,\\
&\tau(0)=s,\\
&\tau(1)=g,
\end{aligned}
\label{eq:mcr_definition}
\end{equation}
where $\tau$ is a continuous collision-free path with respect to all obstacles not contained in $S$.
\label{definition1}
\end{definition}

$S^\star$ is therefore an obstacle set of minimum cardinality whose removal admits a path from $s$ to $g$.
If such a path exists without removing any obstacle, then $S^\star=\emptyset$. 
When several minimum-cardinality sets exist, $S^\star$ denotes any
one of them.

The \textsf{MCR} problem can be represented as a graph-search problem~\cite{erickson2013AAAI, hauser2014IJRR, gorbenko2015AIP, eiben2018ICALP}. 
The overall obstacle arrangement partitions $\mathcal{C}$ into connected cells such that the set of intersecting obstacles is constant within each cell.
Each cell is represented by a vertex $v\in V$, and two vertices are connected by an edge $e\in E$ when their corresponding cells are adjacent.
The start and goal configurations are represented by the designated vertices $s$ and $g$, therefore yielding an \textsf{MCR} graph $G=(V,E)$.

\begin{figure*}[t]
\subfloat[]{\includegraphics[trim=52 80 40 72,clip,scale=0.404]{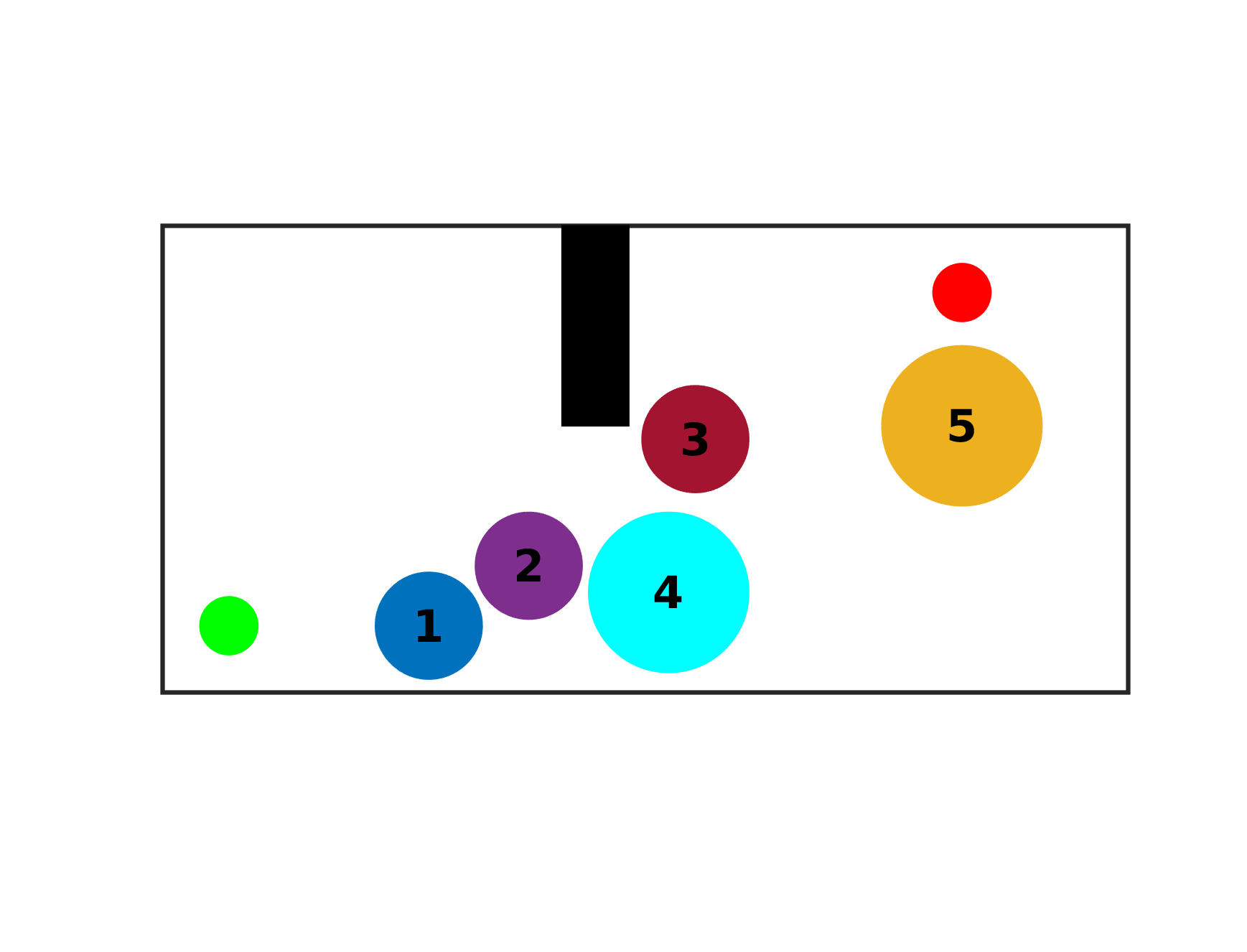}\label{fig:prelim1}}\hspace{0.1cm}
\subfloat[]{\includegraphics[trim=52 84 40 72,clip,scale=0.425]{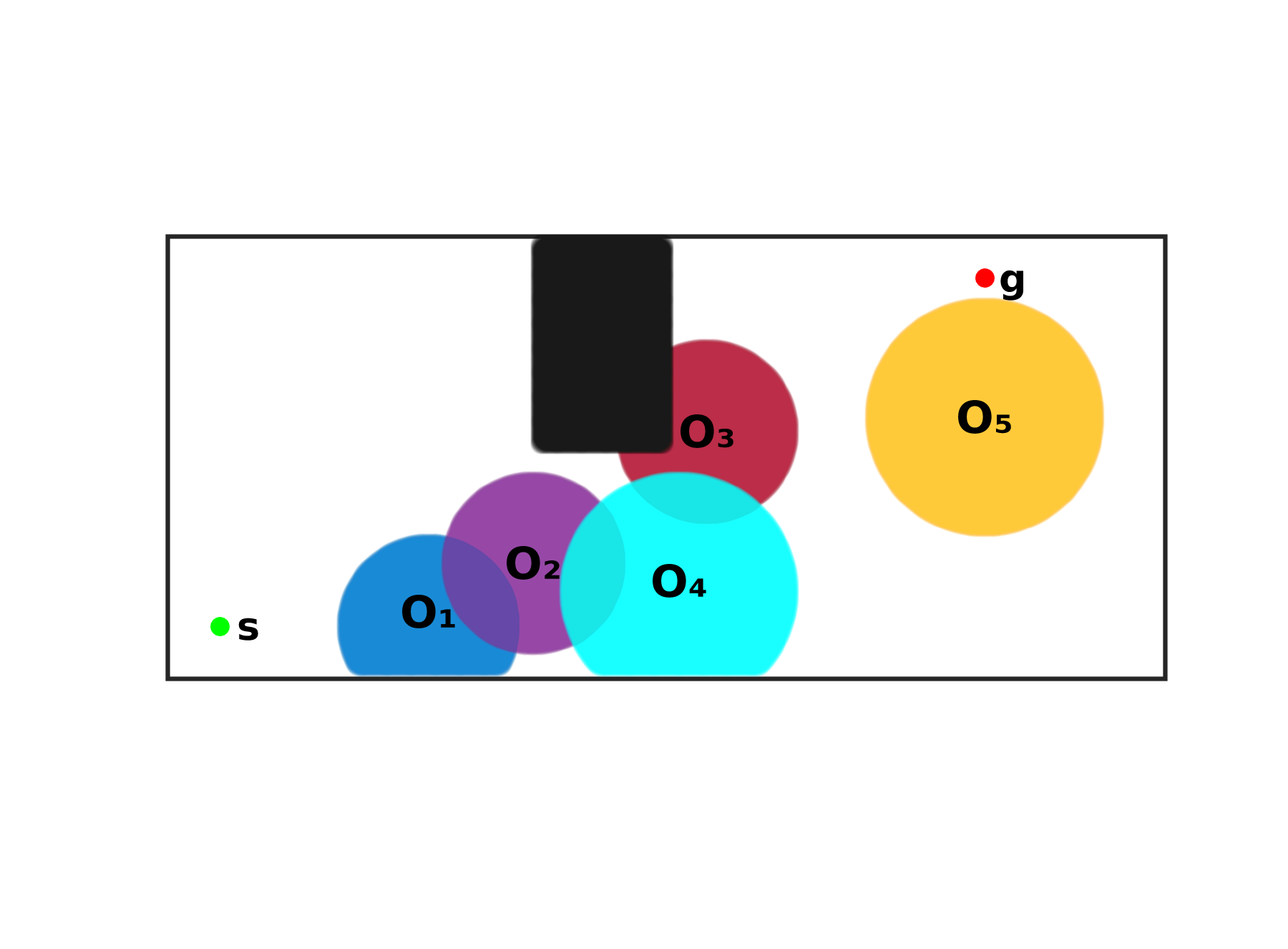}\label{fig:prelim2}}\hspace{0.1cm}
\subfloat[]{\resizebox{0.3\textwidth}{!}{\begin{tikzpicture}[node distance={12mm}, thick, main/.style = {draw, circle}] 
\node[main] (1) [color=green] {$s$}; 
\node[main] (2) [right of=1, xshift=0.5cm, yshift=-2cm] {$1$}; 
\node[main] (3) [right of=2, xshift=0.5cm, yshift=1.2cm] {$1,2$}; 
\node[main] (4) [right of=3, xshift=0.5cm, yshift=1.3cm] {$2$}; 
\node[main] (5) [below of=4, xshift=0.7cm, yshift=0.5cm] {$2,4$}; 
\node[main] (6) [below left of=5, yshift=-1cm, xshift=-0.5cm] {$4$}; 
\node[main] (7) [right of=6, xshift=1cm, yshift=0.2cm] {$3,4$}; 
\node[main] (8) [right of=4, xshift=2cm, yshift=-0.7cm] {$3$}; 
\node[main] (9) [below of=7, xshift=0.5cm] {$5$};
\node[main] (10) [color=red, right of=7, xshift=0.5cm, yshift=-0.9cm] {$g$};
\draw (1) -- (2);
\draw (1) -- (3);
\draw (1) -- (4); 
\draw (1) -- (6);
\draw (1) node[below right=1pt and 5pt]{} to[bend left=30] (8);
\draw (2) -- (3);
\draw (2) node[below right=1pt and 5pt]{} to[bend left=25] (4);
\draw (2) -- (6);
\draw (2) -- (8);
\draw (3) -- (4);
\draw (4) -- (6);
\draw (4) -- (5);
\draw (4) -- (8);
\draw (5) -- (6);
\draw (6) -- (7);
\draw (6) -- (8);
\draw (6) -- (9);
\draw (6) -- (10);
\draw (7) -- (8);
\draw (8) -- (9);
\draw (8) -- (10);
\draw (9) -- (10);
\end{tikzpicture}}
\label{fig:prelim3}}
\caption{
Construction of an \textsf{MCR} graph.
(a) A circular robot moves from the \textit{green} start configuration to the \textit{red} goal configuration among a \textit{black} fixed obstacle and five movable obstacles.
(b) The corresponding configuration space; colors associate workspace objects with their configuration space obstacle regions $o_1, \ldots,o_5$.
(c) The resulting graph, whose vertices represent connected cells and whose labels identify the corresponding obstacle sets $I(v)$.
}
\label{fig:prelim}
\end{figure*}

For each vertex $v$, let
\begin{equation}
I(v)\subseteq\mathcal{O}
\end{equation}
denote its obstacle label, that is, the set of obstacles intersecting the cell represented by $v$.
In particular, $I(v)=\emptyset$ for a free-space cell. 
Figure~\ref{fig:prelim} illustrates this construction for a circular robot moving among five movable obstacles. 
In particular, Figure~\ref{fig:prelim1} shows the workspace, Figure~\ref{fig:prelim2} the corresponding configuration space, and Figure~\ref{fig:prelim3} the resulting \textsf{MCR} graph.

Let
\[
P_G = (v_0 = s, v_1, \ldots, v_k = g)
\]
be a path on $G$.
Its \emph{removable set} is
\begin{equation}
R(P_G) = \bigcup_{j=0}^{k} I(v_j).
\label{eq:removable_set}
\end{equation}
The union in~\eqref{eq:removable_set} is essential, as an obstacle encountered multiple times along a path contributes only once to its cost.
The graph formulation of \textsf{MCR} is therefore
\begin{equation}
P_G^\star\in
\arg\min_{P_G\in\mathcal{P}_G(s,g)} |R(P_G)|,
\qquad
S^\star=R(P_G^\star),
\label{eq:graph_mcr}
\end{equation}
where $\mathcal{P}_G(s,g)$ is the set of graph paths connecting $s$ and $g$. 
For a partial path $P_G$ from $s$ to a vertex $v$, we denote its removable set as $R_P(v)$.
Similarly, $R_P(v'\rightarrow v)$ denotes the removable set associated with a path segment from $v'$ to $v$. 
The path subscript is omitted whenever the relevant path is unambiguous.

Hereafter, \textsf{MCR} refers to the graph-search formulation.
We assume that $G$ is available, and its construction from a robot model and workspace is discussed in Section~\ref{subsec:examples}. 
The graph labels represent configuration space obstacles induced by movable workspace
objects.
The present formulation determines \emph{which} objects should be removed, and assigns unit cost to every removal.
It does not determine how or where an object should be relocated, nor whether one object is more difficult to manipulate than another.

An exact solution can be computed using a best-first search whose states are pairs $(v, R)$, where $v$ is the current vertex and $R$ is the removable set accumulated along the corresponding path.
Expanding a successor $v'$ produces the state
\[
\bigl(v',R\cup I(v')\bigr).
\]
Expanding states in non-decreasing order of $|R|$ yields a complete and optimal search~\cite{hauser2014IJRR}.
However, multiple removal sets may have to be retained at the same
vertex.
In particular, two sets of equal or different cardinality can lead to different future costs because an obstacle already contained in $R$ can be encountered again
without increasing $|R|$. 
Consequently, retaining only the lowest-cardinality state at each vertex, as in a conventional greedy search, can discard a path that later yields a better
global solution.

\begin{figure}[]
\subfloat[]{\includegraphics[scale=0.4]{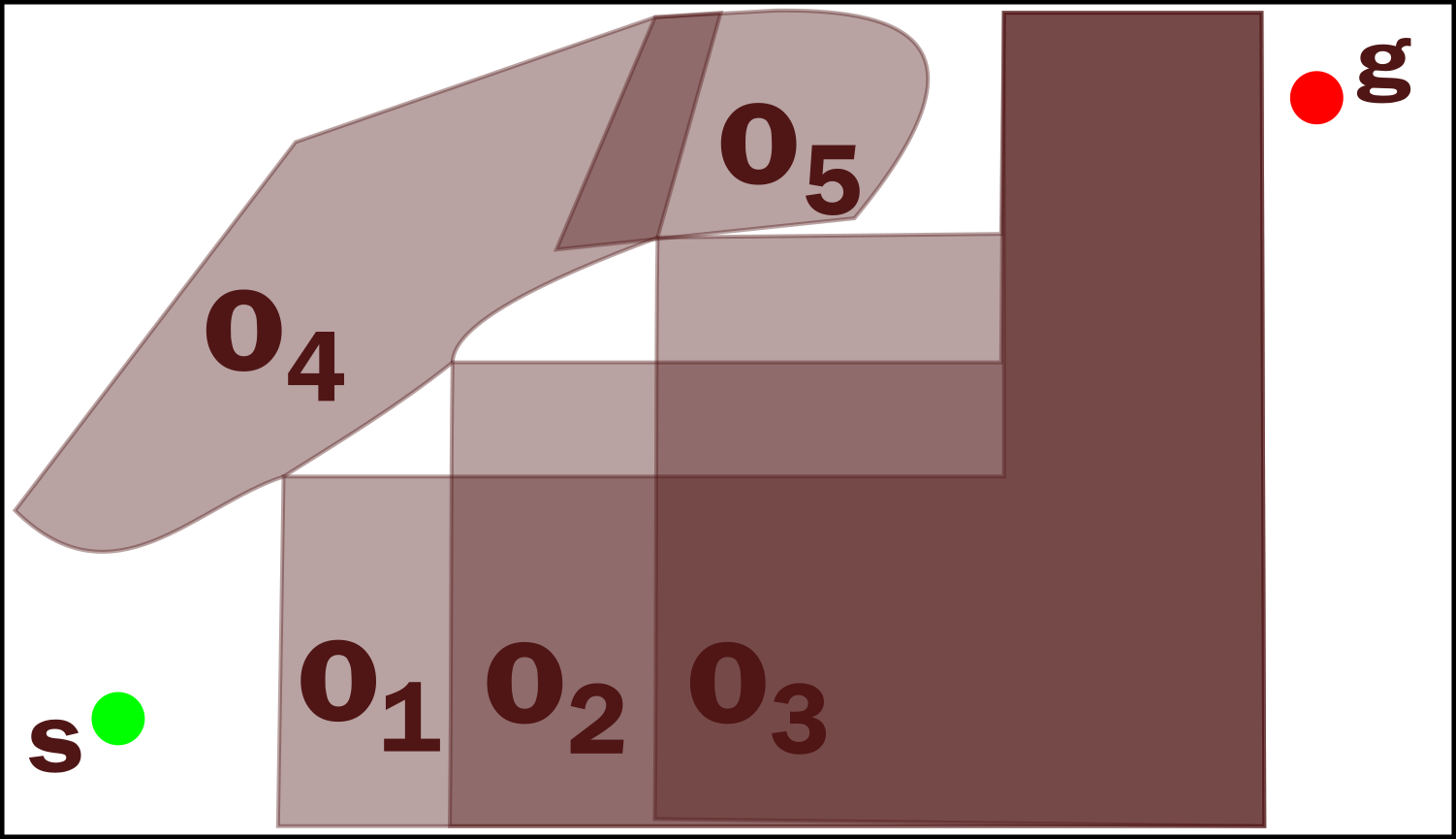}\label{fig:C1}}\hspace{0.1cm}
\subfloat[]{\resizebox{0.4\textwidth}{!}{\begin{tikzpicture}[node distance={12mm}, thick, main/.style = {draw, circle}] 
\node[main] (1) [color=green] {$s$}; 
\node[main] (2) [above of=1, yshift=1cm] {$4$}; 
\node[main] (3) [below of=2, xshift=1cm] {$1$}; 
\node[main] (4) [right of=3, xshift=0.5cm] {$1,2$}; 
\node[main] (5) [above of=3, xshift=0.5cm] {$2$}; 
\node[main] (6) [above right of=2, yshift=1.2cm] {$4,5$}; 
\node[main] (7) [right of=6, xshift=3cm, yshift=-0.8cm] {$5$}; 
\node[main] (8) [below right of=6, yshift=-0.3cm] {$3$}; 
\node[main] (9) [right of=4, xshift=0.5cm] {$1,2,3$};
\node[main] (10) [below of=8, yshift=0.5cm, xshift=2cm] {$2,3$};
\node[main] (11) [color=red, right of=9, xshift=0.5cm] {$g$};
\draw (1) -- (2); 
\draw (1) -- (3); 
\draw (2) -- (3);
\draw (2) -- (5);
\draw (2) -- (6);
\draw (2) -- (8);
\draw (3) -- (4);
\draw (4) -- (5);
\draw (4) -- (9);
\draw (5) -- (8);
\draw (6) -- (7);
\draw (7) -- (9);
\draw (7) -- (8);
\draw (8) -- (10);
\draw (8) -- (9);
\draw (9) -- (10);
\draw (9) -- (11);
\end{tikzpicture}}\label{fig:C2}}\hfill
\caption{An \textsf{MCR} instance illustrating the path dependence of removable sets.
(a) Five overlapping obstacles in a two-dimensional configuration space.
(b) The corresponding undirected \textsf{MCR} graph.
Each vertex represents a connected cell, and its label lists the obstacles
intersecting that cell.
}
\label{fig:C}
\end{figure}

An example is shown in Figure~\ref{fig:C}.
In particular, Figure~\ref{fig:C2} illustrates this lack of scalar optimal substructure.
Let us consider the vertex $v$ labeled by $I(v)=\{o_3\}$.
A minimum-cardinality path from $s$ to $v$ has removal set $\{o_3, o_4\}$, whereas a minimum-cardinality path from $v$ to $g$ has removal set $\{o_1, o_2, o_3\}$. Combining these locally optimal subpaths therefore produces the set
$\{o_1, o_2, o_3, o_4\}$.
Nevertheless, an $s$--$g$ path through $v$ exists with removal set $\{o_1, o_2, o_3\}$.
Therefore, a globally optimal path may contain a prefix that is not locally
optimal in terms of removal-set cardinality.

\section{Graph-Based Approximation Heuristics}
\label{sec:polynomial}
\subsection{Rationale}
\label{subsec:rationale}

We propose a heuristic pipeline that first generates feasible obstacle-removal sets and then refines them through bounded-length searches on reduced \textsf{MCR} graphs. For a candidate set $C\subseteq\mathcal O$, let us define
\begin{equation}
V_C = \{s,g\}\cup \{v\in V\mid I(v)\subseteq C\}, \qquad G_C=G[V_C],
\label{eq:reduced_graph}
\end{equation}
where $G[V_C]$ is the subgraph of $G$ induced by $V_C$. 
The candidate $C$ is \emph{feasible} if $s$ and $g$ are connected in $G_C$.
Equivalently, removing the obstacles in $C$ admits a collision-free path between the start and goal configurations.

The first feasible-set generator, denoted by \textsf{GFS-1}, performs bidirectional exploration from $s$ and $g$.
Let $\widehat G$ denote the current working copy of the \textsf{MCR} graph, and let $\widehat I(v)$ denote the current obstacle label of vertex $v$ in $\widehat G$.
Initially, it holds that $\widehat G=G$ and $\widehat I(v)=I(v)$ for every $v\in V$.
Removing obstacle labels modifies $\widehat I$, whereas the original labels $I(v)$ remain unchanged.

Let $F^s$ and $F^g$ denote the current search frontiers, and let $X^s$ and $X^g$ contain the vertices visited by the respective searches.
The operation $\textsf{Expand}(\widehat G,F,X)$ returns the unvisited neighbors of the current frontier $F$ in $\widehat G$ and adds them to the visited set $X$.

At depth $l$, the algorithm collects the current obstacle labels represented in the two frontiers:
\begin{equation}
IO^s=\bigcup_{v\in F^s}\widehat I(v),
\qquad
IO^g=\bigcup_{v\in F^g}\widehat I(v).
\label{eq:frontier_labels}
\end{equation}
Labels occurring in both frontiers are added to the current
candidate set.
The operation $\textsf{RemoveAndContract}(\widehat G,RO)$ removes every label
in $RO$ from the working labels $\widehat I(v)$, and contracts vertices whose labels consequently become empty. 
This contraction preserves free-space connectivity.
Reinitializing $\widehat G\gets G$ also restores
$\widehat I(v)\gets I(v)$ for every $v\in V$.

Whenever the accumulated candidate makes $s$ and $g$ connected, the pair $(C,l)$ is added to the candidate collection $RS$.
After a feasible candidate $C$ is recorded, the graph and search frontiers are reinitialized, and $U\gets U\cup C$ stores all obstacle labels contained in previously generated candidates.
Labels in $U$ are excluded from subsequent candidates.
Consequently, the nonempty candidate sets generated by \textsf{GFS-1} are disjoint, and their number is at most $n = |\mathcal O|$.

\begin{algorithm}[t]
\caption{\textsf{GenerateFeasibleSets-1} (\textsf{GFS-1})}
\label{algo:GS}
\begin{algorithmic}[1]
\Require{$\mathcal O$, $s$, $g$, $G=(V,E)$, and $I(v)$
for every $v\in V$}
\State $RS\gets\emptyset$, $U\gets\emptyset$
\If{\textsf{Feasible}$(G,\emptyset)$}
    \State \Return $\{(\emptyset,0)\}$
\EndIf
\State $\mathit{active}\gets\texttt{true}$
\While{$\mathit{active}$ and $U\neq\mathcal O$}
    \State $\widehat G\gets G$
    \State $F^s\gets\{s\}$, $F^g\gets\{g\}$
    \State $X^s\gets\{s\}$, $X^g\gets\{g\}$
    \State $C\gets\emptyset$, $l\gets0$
    \State $\mathit{found}\gets\texttt{false}$
    \While{$F^s\neq\emptyset$ and $F^g\neq\emptyset$}
        \State $(F^s,X^s)\gets\textsf{Expand}(\widehat G,F^s,X^s)$
        \State $(F^g,X^g)\gets\textsf{Expand}(\widehat G,F^g,X^g)$
        \State $l\gets l+1$
        \State $IO^s\gets\displaystyle\bigcup_{v\in F^s}\widehat I(v)$
        \State $IO^g\gets\displaystyle\bigcup_{v\in F^g}\widehat I(v)$
        \State $RO\gets(IO^s\cap IO^g)\setminus U$
        \State $C\gets C\cup RO$
        \State $\widehat G\gets
        \textsf{RemoveAndContract}(\widehat G,RO)$
        \If{$C\neq\emptyset$ and \textsf{Feasible}$(G,C)$}
            \State $RS\gets RS\cup\{(C,l)\}$
            \State $U\gets U\cup C$
            \State $\mathit{found}\gets\texttt{true}$
            \State \textbf{break}
        \EndIf
    \EndWhile
    \If{$\neg\mathit{found}$}
        \State $\mathit{active}\gets\texttt{false}$
    \EndIf
\EndWhile
\State \Return $RS$
\end{algorithmic}
\end{algorithm}

\subsection{Reduced-Graph Candidate Refinement}
\label{subsec:refinement}

A candidate produced by \textsf{GFS-1} is feasible but need not be minimal.
Moreover, the smallest generated candidate is not necessarily the one containing the best refined solution.
Therefore, every pair $(C,l)\in RS$ is examined independently.

For each candidate, the subroutine \textsf{ReducedGraph} constructs the graph $G_C$ in~\eqref{eq:reduced_graph}.
Every $s$--$g$ path in $G_C$ has a removable set contained in $C$.
The refinement stage searches paths of selected lengths and
returns the one whose removable set has minimum cardinality.

The search depth $l$ provides an estimate of the relevant path
length.
Let
\begin{equation}
r_C=
\max_{A\subseteq C}
\left|\{v\in V_C\mid I(v)=A\}\right|
\end{equation}
be the maximum multiplicity of an obstacle label in $G_C$. 
For an ordinary bidirectional meeting, the searched path lengths are
\begin{equation}
\mathcal L(C,l)=
\{2l, 2l+1, \ldots, 2l+r_C-1\}.
\label{eq:admissible_lengths}
\end{equation}
If $l=0$, then $\mathcal L(C,0)=\{1\}$. 
When the two search frontiers are adjacent to the opposite terminal vertex, the
lengths $l+1$ and $l+2$ are also included. 
Path length is measured in graph edges.

For a fixed length $L$, \textsf{FixedLengthSearch} explores $s$--$g$ paths of exactly $L$ edges and returns a minimum-cardinality removable set among them. 
At the same vertex and search depth, a state with removable set $R_2$ can be pruned when another state has removable set $R_1 \subseteq R_2$. 
Cardinality alone is not a valid dominance condition.

\begin{algorithm}[t]
\caption{\textsf{Refine}}
\label{algo:GS2}
\begin{algorithmic}[1]
\Require{$G=(V,E)$, $s$, $g$, $I(v)$ for every $v\in V$,
and $RS$}
\State $S_{\mathrm{best}}\gets\bot$
\ForAll{$(C,l)\in RS$}
    \State $G_C\gets\textsf{ReducedGraph}(G,C)$
    \State $\mathcal L\gets
    \textsf{AdmissibleLengths}(G_C,l)$
    \ForAll{$L\in\mathcal L$}
        \State $S\gets
        \textsf{FixedLengthSearch}(G_C,s,g,L)$
        \If{$S\neq\bot$ and
        $(S_{\mathrm{best}}=\bot$ or
        $|S|<|S_{\mathrm{best}}|)$}
            \State $S_{\mathrm{best}}\gets S$
        \EndIf
    \EndFor
\EndFor
\State \Return $S_{\mathrm{best}}$
\end{algorithmic}
\end{algorithm}

\begin{figure}[]
\subfloat[]{
\includegraphics[scale=0.5]{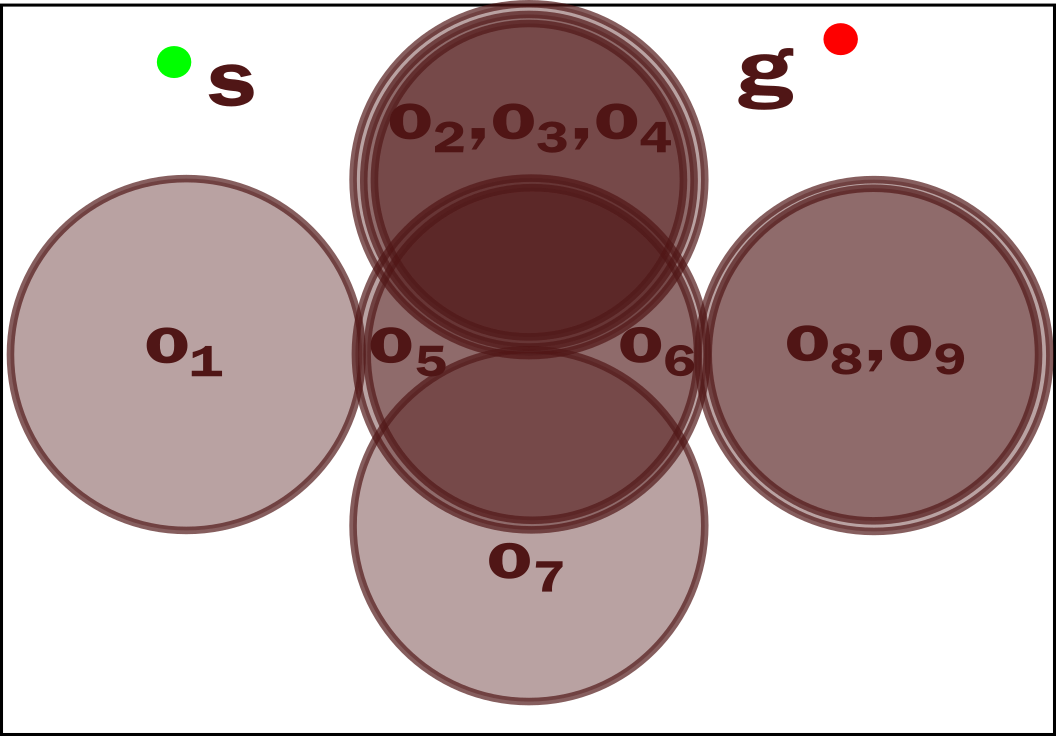}
\label{fig:twosol}
}
\hfill
\subfloat[]{
\resizebox{0.5\textwidth}{!}
{
\begin{tikzpicture}[node distance={12mm}, thick, main/.style = {draw, circle}] 
\node[main] (1) [color=green] {$s$}; 
\node[main] (2) [right of=1] {$1$}; 
\node[main] (3) [above right of=2, xshift=1cm, yshift=3cm] {$2,3,4$}; 
\node[main] (4) [right of=2, xshift=0.5cm, yshift=0.2cm] {$5,6$}; 
\node[main] (5) [above of=4, xshift=0.7cm, yshift=0.8cm] {$2,3,4,5,6$}; 
\node[main] (6) [above right of=4, xshift=2.7cm] {$5,6$}; 
\node[main] (7) [below of=5, yshift=-1.4cm,xshift=1.2cm] {$5,6,7$}; 
\node[main] (8) [below of=7, xshift=0.5cm,yshift=-0.3cm] {$7$}; 
\node[main] (9) [color=red, above right of=6, xshift=0.5cm, yshift=-0.6cm] {$g$};
\node[main] (10) [below of=9, yshift=0.5cm, xshift=2cm] {$8,9$};
\draw (1) -- (2); 
\draw (1) -- (3); 
\draw (2) -- (3);
\draw (2) -- (4);
\draw (2) -- (8);
\draw (2) node[below right=1pt and 5pt]{} to[bend right=90] (10);
\draw (1) node[below right=1pt and 5pt]{} to[bend right=40] (4);
\draw (3) -- (9);
\draw (3) node[below right=1pt and 5pt]{} to[bend right=20] (4);
\draw (3) -- (5);
\draw (3) node[below right=1pt and 5pt]{} to[bend right=-5] (6);
\draw (3) node[below right=1pt and 5pt]{} to[bend right=-20] (10);
\draw (4) -- (5);
\draw (4) -- (6);
\draw (4) -- (7);
\draw (4) -- (8);
\draw (4) node[below right=1pt and 12pt]{} to[bend left=42] (10);
\draw (5) -- (6);
\draw (5) -- (7);
\draw (6) -- (7);
\draw (6) -- (8);
\draw (6) -- (9);
\draw (6) -- (10);
\draw (7) -- (8);
\draw (8) -- (10);
\draw (9) -- (10);
\end{tikzpicture}}
\label{fig:graph}}
\caption{
An \textsf{MCR} instance adapted from \cite{hauser2014IJRR}.
(a) Nine circular obstacles in a two-dimensional configuration space.
(b) The corresponding \textsf{MCR} graph.
The example shows that the smallest candidate generated by \textsf{GFS-1} need not contain the optimal refined solution.}
\label{fig:hauser}
\end{figure} 

Figure~\ref{fig:hauser} illustrates why candidate refinement is
necessary.
For this instance, \textsf{GFS-1} generates
\[
C_1=\{o_2, o_3, o_4, o_5, o_6\},
\qquad
C_2=\{o_1, o_7, o_8, o_9\}.
\]
Although $|C_2|<|C_1|$, the optimal solution $S^\star=\{o_5, o_6\}$ is contained in $C_1$.
Searching the reduced graph associated with $C_1$ can therefore recover a solution that is smaller than either original candidate.

However, candidate containment is necessary but not sufficient for recovering an \textit{optimum}.
Even if some $C\in RS$ contains an optimal removal set, the corresponding optimal path may have a length outside $\mathcal L(C,l)$.
Consequently, the proposed pipeline is heuristic and does not provide a worst-case
approximation guarantee.
For every returned set $S$, feasibility guarantees only that
\[
|S|\geq |S^\star|;
\]
it does not imply the set-containment relation
$S\supseteq S^\star$.

\subsection{Extended Feasible-Set Generation}
\label{subsec:gfs2}

The refinement stage can recover an optimal solution only if at least one generated candidate contains the obstacle set of an optimal path and the associated path length is searched.
However, \textsf{GFS-1} may generate candidates that do not contain any complete optimal removal set.

\begin{figure}[t!]
\centering
\subfloat[]{%
    \includegraphics[
        width=0.48\linewidth
    ]{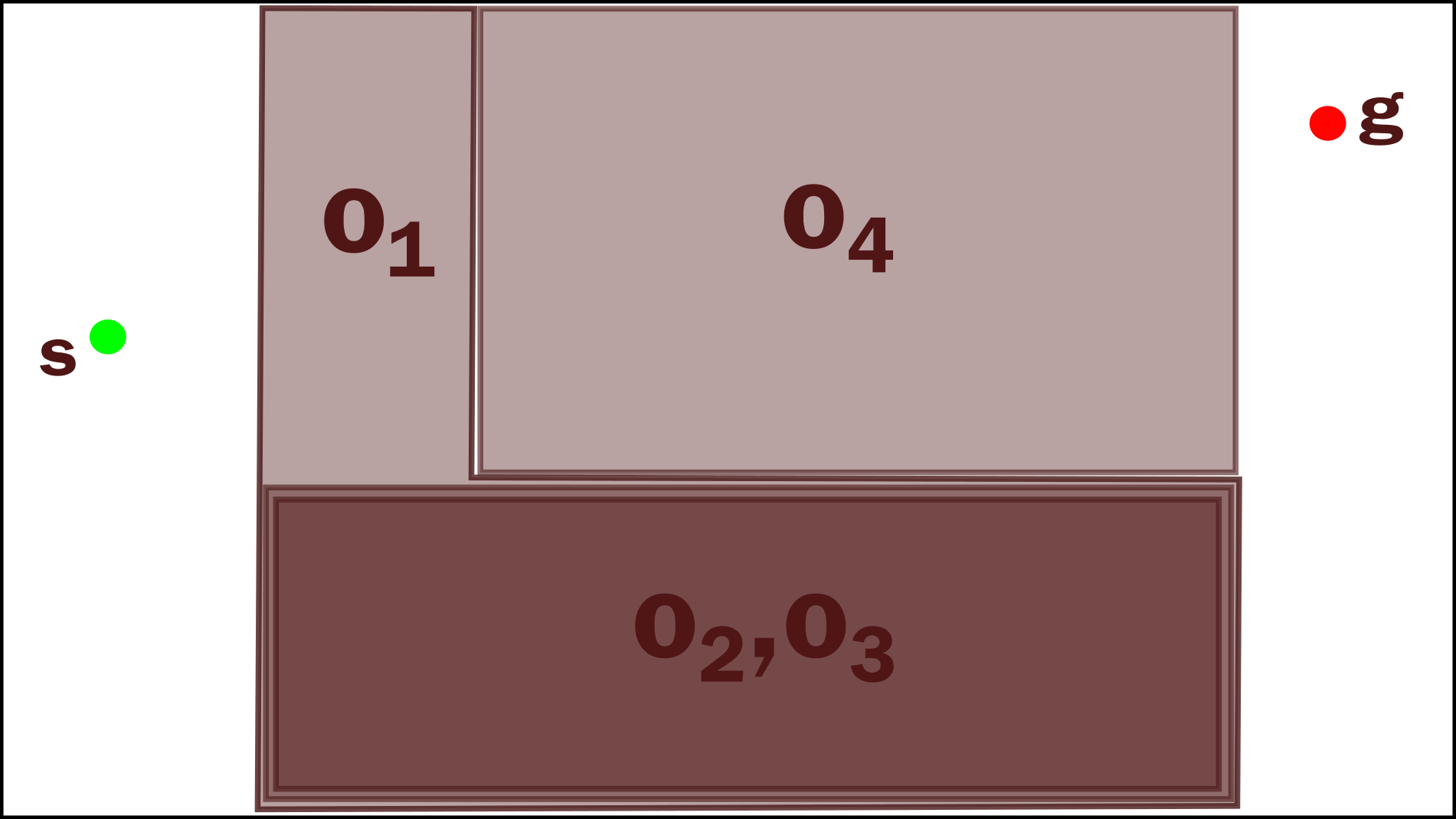}%
    \label{fig:counter}%
}
\hfill
\subfloat[]{%
    \resizebox{0.48\linewidth}{!}{%
        \begin{tikzpicture}[
            node distance={12mm},
            thick,
            main/.style={draw,circle}
        ]
        \node[main] (1) [color=green] {$s$};
        \node[main] (2) [right of=1] {$1$};
        \node[main] (3) [right of=2] {$4$};
        \node[main] (4) [below right of=1,
            xshift=0.5cm,yshift=-0.5cm] {$1,2,3$};
        \node[main] (5) [color=red,
            below right of=3] {$g$};

        \draw (1) -- (2);
        \draw (1) -- (4);
        \draw (2) -- (3);
        \draw (2) -- (4);
        \draw (3) -- (5);
        \draw (3) -- (4);
        \draw (4) -- (5);
        \end{tikzpicture}%
    }%
    \label{fig:counter_graph}%
}
\caption{
An \textsf{MCR} instance motivating \textsf{GFS-2}.
(a) A two-dimensional configuration space containing the L-shaped obstacle $o_1$ and rectangular obstacles $o_2, o_3, o_4$.
(b) The corresponding \textsf{MCR} graph.
}
\label{fig:counterex}
\end{figure}

Let us consider the instance in Figure~\ref{fig:counterex}.
\textsf{GFS-1} generates $C_1=\{o_1, o_2, o_3\}$, whereas an optimal solution is
$S^\star=\{o_1, o_4\}$. 
Although the candidate and the optimal solutions share $o_1$, $S^\star\nsubseteq C_1$, and refinement restricted to $G_{C_1}$ cannot recover the optimum.
To address this case, \textsf{GFS-2} does not exclude labels that appeared in previously generated candidates.
Moreover, after finding a feasible candidate, it reinitializes the working graph but retains the current bidirectional frontiers and visited sets. 
The search, therefore, continues at increasing depths and can generate larger candidates containing combinations that \textsf{GFS-1} separates across restarts.

\begin{algorithm}[t]
\caption{\textsf{GenerateFeasibleSets-2} (\textsf{GFS-2})}
\label{algo:GS3}
\begin{algorithmic}[1]
\Require{$\mathcal O$, $s$, $g$, $G=(V,E)$, and $I(v)$
for every $v\in V$}
\State $RS\gets\emptyset$
\If{\textsf{Feasible}$(G,\emptyset)$}
    \State \Return $\{(\emptyset,0)\}$
\EndIf
\State $\widehat G\gets G$
\State $F^s\gets\{s\}$, $F^g\gets\{g\}$
\State $X^s\gets\{s\}$, $X^g\gets\{g\}$
\State $C\gets\emptyset$, $l\gets0$
\While{$F^s\neq\emptyset$ and $F^g\neq\emptyset$}
    \State $(F^s,X^s)\gets\textsf{Expand}(\widehat G,F^s,X^s)$
    \State $(F^g,X^g)\gets\textsf{Expand}(\widehat G,F^g,X^g)$
    \State $l\gets l+1$
    \State $IO^s\gets\displaystyle\bigcup_{v\in F^s}\widehat I(v)$
    \State $IO^g\gets\displaystyle\bigcup_{v\in F^g}\widehat I(v)$
    \State $RO\gets IO^s\cap IO^g$
    \State $C\gets C\cup RO$
    \State $\widehat G\gets \textsf{RemoveAndContract}(\widehat G,RO)$
    \If{$C\neq\emptyset \land \textsf{Feasible}(G,C)$}
        \State $RS\gets RS\cup\{(C,l)\}$
        \State $C\gets\emptyset$
        \State $\widehat G\gets G$
        \Comment{retain $F^s, F^g, X^s, X^g$}
    \EndIf
\EndWhile
\State \Return $RS$
\end{algorithmic}
\end{algorithm}

For the instance in Fig.~\ref{fig:counterex}, \textsf{GFS-2}
generates
\[
RS=
\left\{
(\{o_1, o_2, o_3\}, 1),
(\{o_1, o_2, o_3, o_4\},2)
\right\}.
\]
The second candidate contains the optimal solution $S^\star=\{o_1, o_4\}$. 
Refinement of its reduced graph therefore recovers the optimal set using a path of length three.

\subsection{Complexity Analysis}
\label{subsec:complexity}

Let $N=|V|$, $M=|E|$, $n=|\mathcal O|$, and let $\Delta$ be the maximum vertex degree encountered during graph contraction.
We formulate complexity directly in terms of the \textsf{MCR} graph because the number of arrangement cells cannot, in general, be bounded by $O(n^2)$ for an arbitrary $d$-dimensional configuration space.

During one bidirectional pass, every vertex and edge is visited at most a constant number of times, giving a traversal cost of $O(N+M)$. 
Contracting a vertex of degree at most $\Delta$ requires at most $O(\Delta^2)$ operations to connect its neighbors.
Thus, contracting at most $N$ vertices during one pass costs $O(N\Delta^2)$. 
These bounds assume that obstacle labels are represented by bit sets and that an inverted index identifies the vertices containing each label.

Let $q_1=|RS|$ be the number of candidates generated by \textsf{GFS-1}. 
Because previously used labels are excluded, its nonempty candidates are disjoint and therefore $q_1\leq n$. 
Its worst-case candidate-generation cost is therefore
\begin{equation}
O\!\left(
q_1\left(N+M+N\Delta^2\right)
\right).
\end{equation}

In \textsf{GFS-2}, the frontiers and visited sets are not reset, so at most $N$ bidirectional depth expansions are performed.
If $q_2$ feasible candidates are generated, graph reinitialization may cause up to $q_2$ separate sequences of contractions. 
A conservative bound is
\begin{equation}
O\!\left(
N+M+q_2N\Delta^2
\right),
\qquad q_2\leq N.
\end{equation}

For a candidate $C$, constructing the induced reduced graph $G_C$ takes $O(N+M)$ time using adjacency lists. 
Let $\Delta_C$ denote its maximum degree. 
Enumerating paths of a fixed length $L$ has worst-case cost $O(\Delta_C^L)$. Therefore, the total refinement cost is bounded by
\begin{equation}
O\!\left(
\sum_{(C,l)\in RS}
\left[
N+M+
\sum_{L\in\mathcal L(C,l)}\Delta_C^L
\right]
\right).
\label{eq:refinement_complexity}
\end{equation}
Subset-based dominance pruning substantially reduces the number of explored states in practice, but it does not change the exponential worst-case dependence on the maximum searched path length. 
Accordingly, the complete pipeline is a heuristic and is not claimed to run in polynomial time in the worst case.
Its empirical computation time is evaluated in Section~\ref{subsec:examples}.

\section{Evaluation}
\label{subsec:examples}

We evaluate the solution quality and computation time of the proposed pipeline against two baselines:
\begin{enumerate}
\item 
the exact best-first \textsf{MCR} search of~\cite{hauser2014IJRR}; and
\item
a greedy search that retains, at each vertex $v$, only the state having the smallest removal-set cardinality; 
specifically, a newly generated state $(v, R')$ is discarded when a previously retained state $(v, R)$ satisfies $|R| \leq |R'|$.
\end{enumerate}
The greedy baseline is computationally inexpensive but does not preserve all path-dependent removal sets and can therefore return a substantially suboptimal solution.
All methods operate on the same \textsf{MCR} graph for each scenario.

The experiments were performed on an Intel Core i7-10510U CPU at 1.80~GHz with 16~GB RAM under Ubuntu 18.04.
All methods were implemented in MATLAB and executed single-threaded.
The reported times measure only graph search.
The construction of the \textsf{MCR} graph is excluded because all methods receive the same precomputed graph as input. 
Each entry reports the mean over 10 repeated runs. 
Since the algorithms are deterministic and the run-to-run variation was negligible at the reported precision, standard deviations are omitted.

\begin{figure}[t]
\centering
\subfloat[$S_1$]{
\includegraphics[trim=50 80 30 70,clip,scale=0.37]
{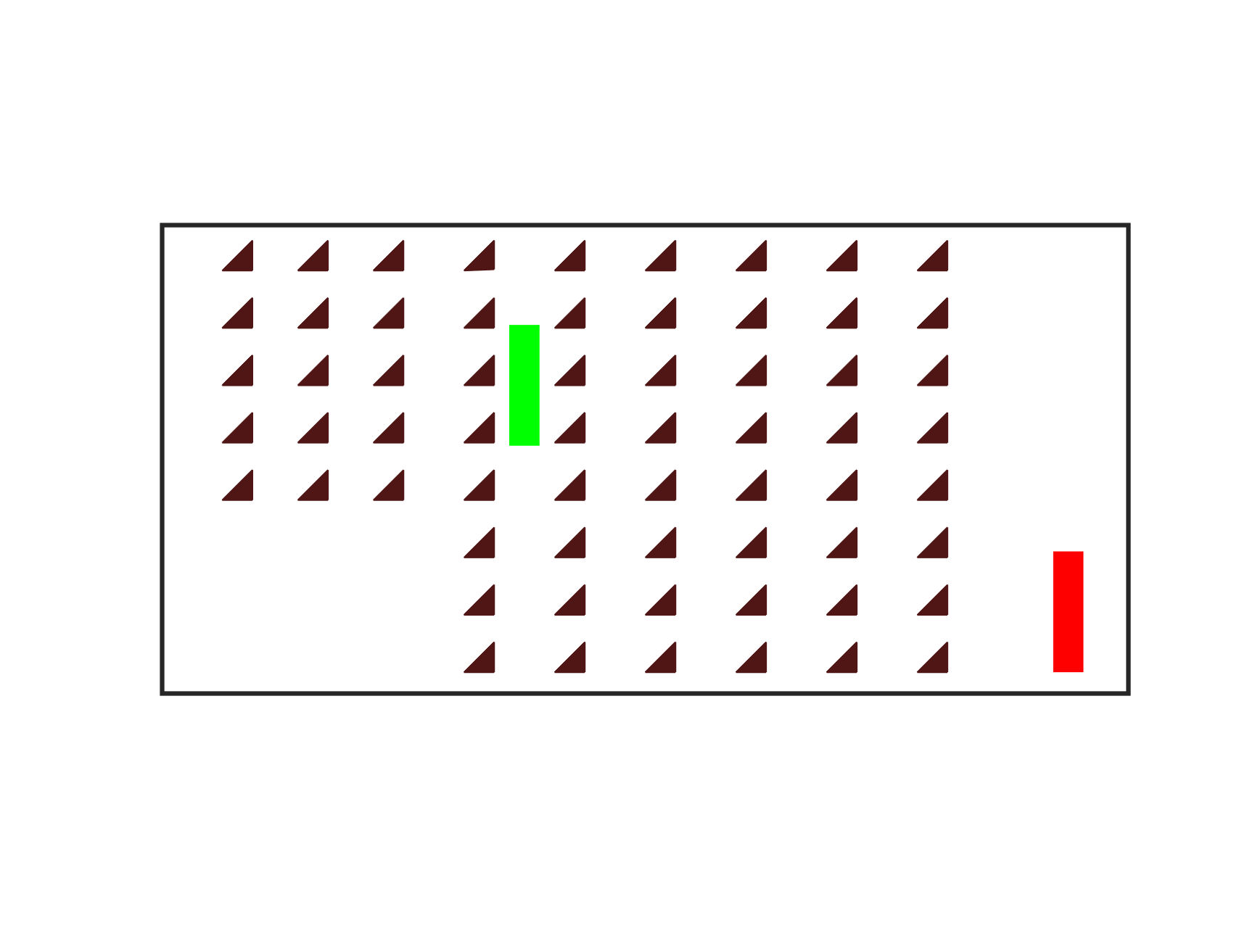}}
\subfloat[\textsf{MCR} path]{
\includegraphics[trim=50 77.5 30 72,clip,scale=0.36]
{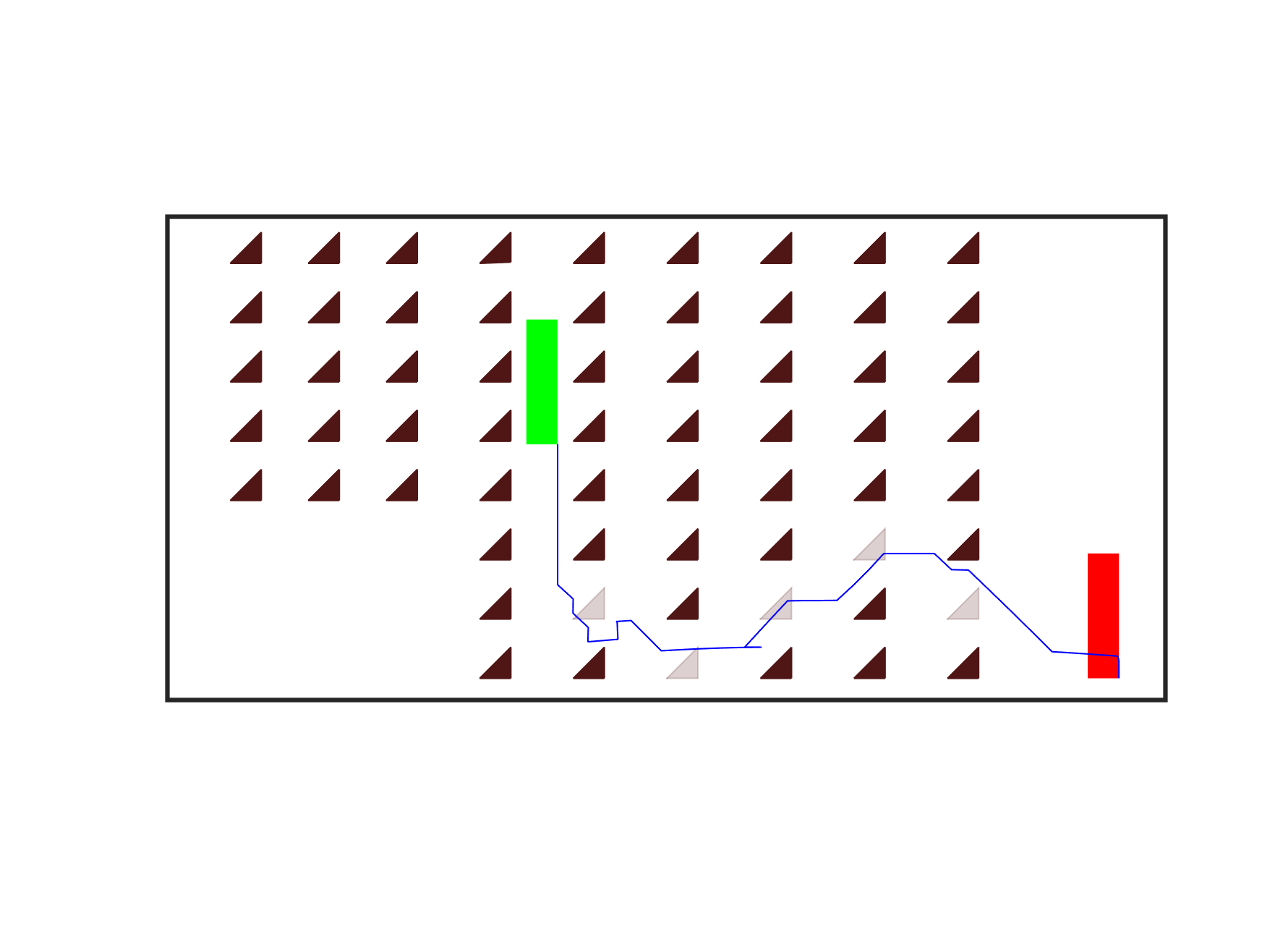}}\\
\subfloat[$S_3$]{
\includegraphics[trim=50 80 30 70,clip,scale=0.371]
{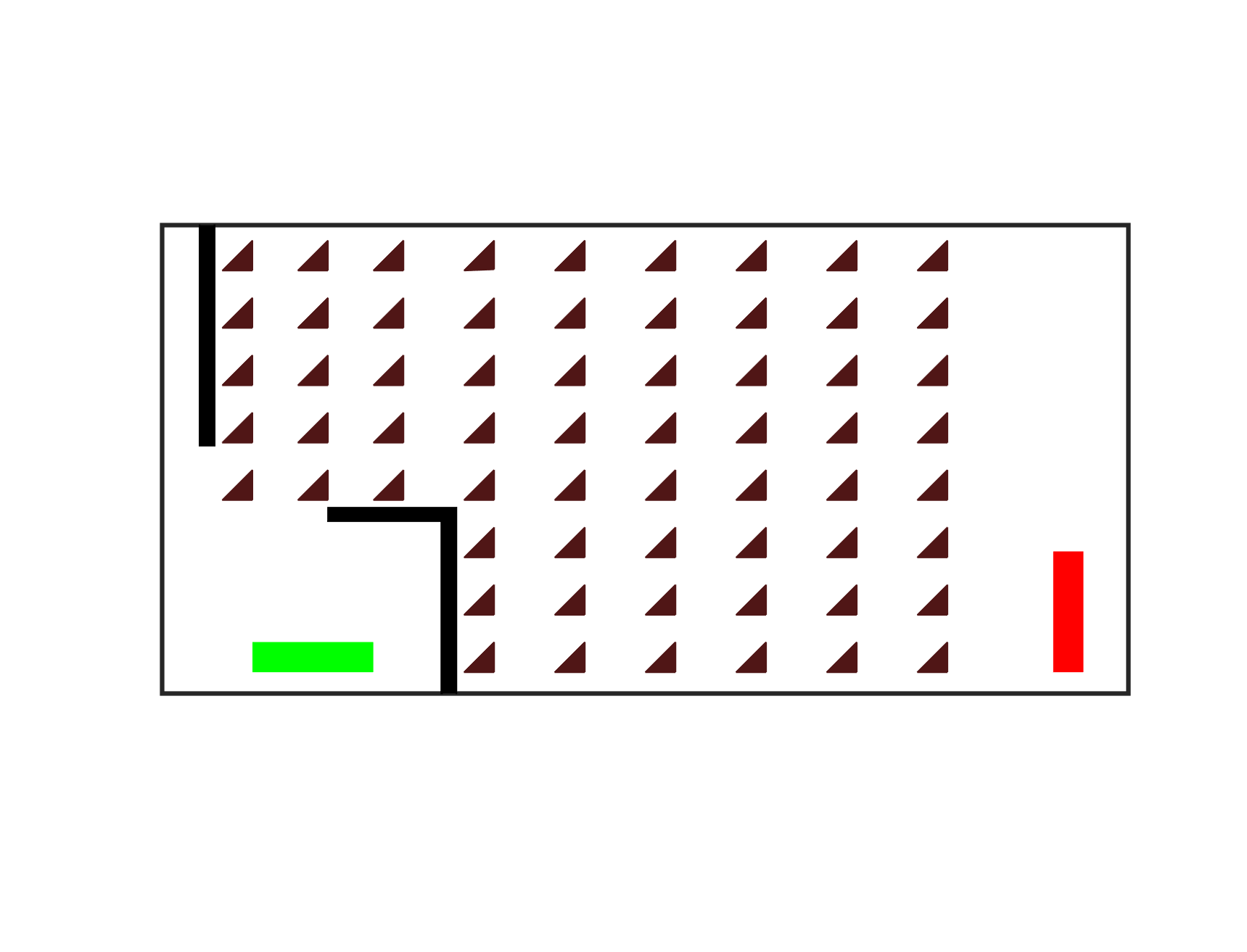}}
\subfloat[\textsf{MCR} path]{
\includegraphics[trim=50 80 30 70,clip,scale=0.367]
{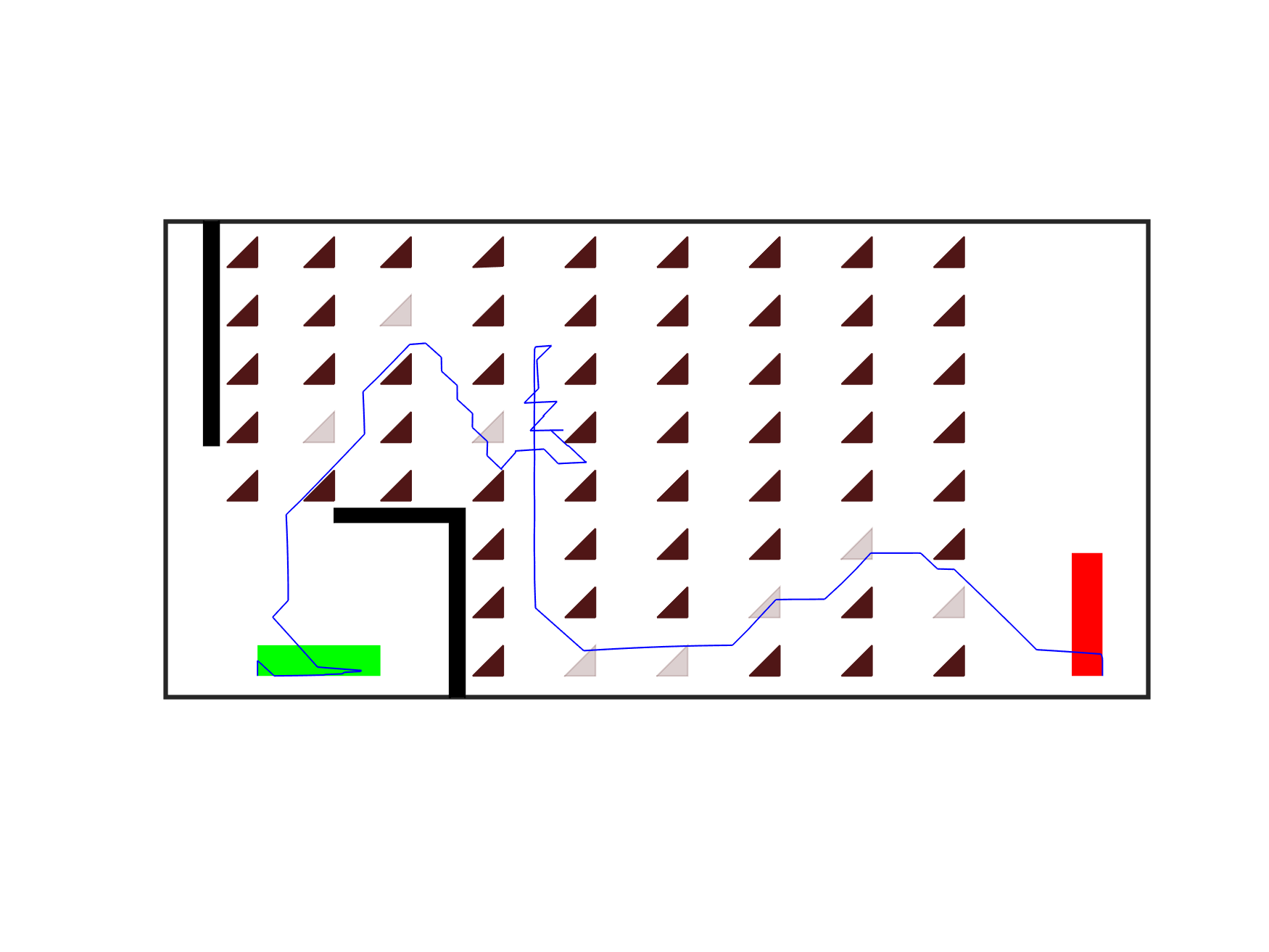}}\\
\subfloat[$S_5$]{
\includegraphics[trim=54 100 35 85,clip,scale=0.37]
{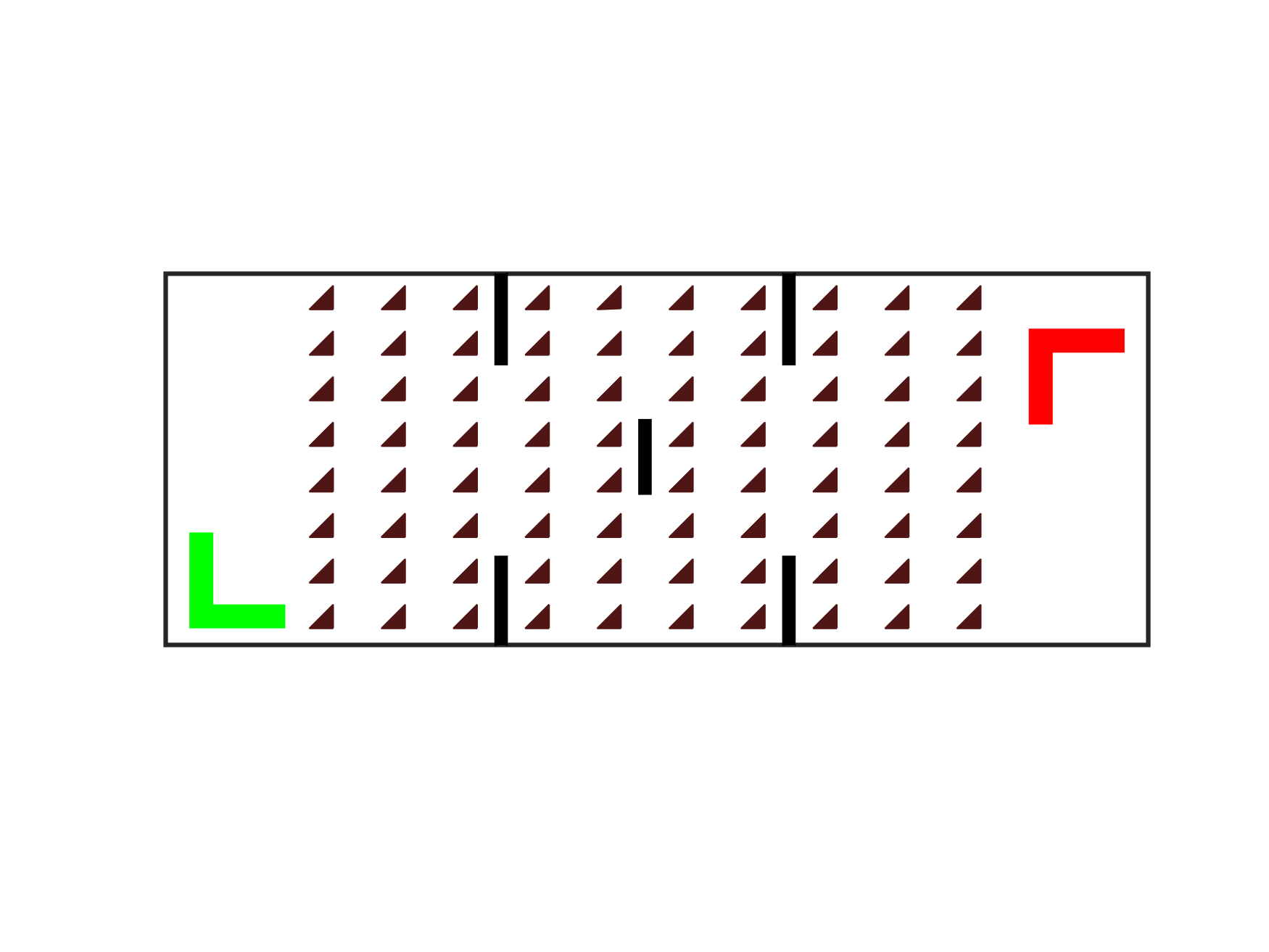}}
\subfloat[\textsf{MCR} path]{
\includegraphics[trim=50 98 40 85,clip,scale=0.36]
{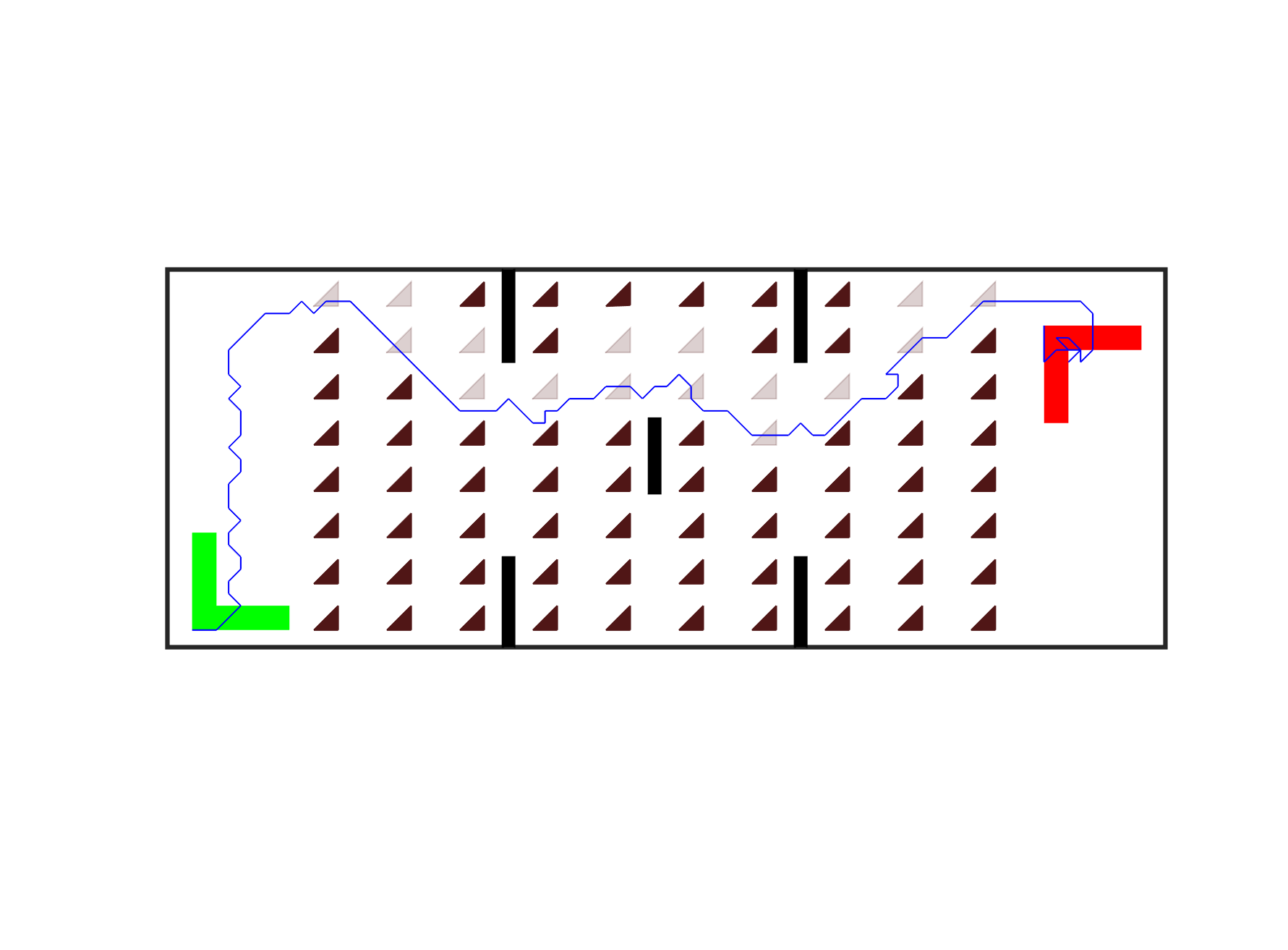}}
\caption{
Representative planar benchmarks.
Left: start configuration in \textit{green}, goal configuration in \textit{red}, and removable triangular obstacles.
Right: paths obtained after removing the obstacles selected by the \textsf{MCR} solution.
Results for all six scenarios are reported in Table~\ref{tab:statistics}.
}
\label{fig:simulations}
\end{figure}

\begin{figure}[]
\subfloat[$S6$: Exact and ours]{
\includegraphics[angle=90, trim=70 15 10 9,clip,scale=0.37]
{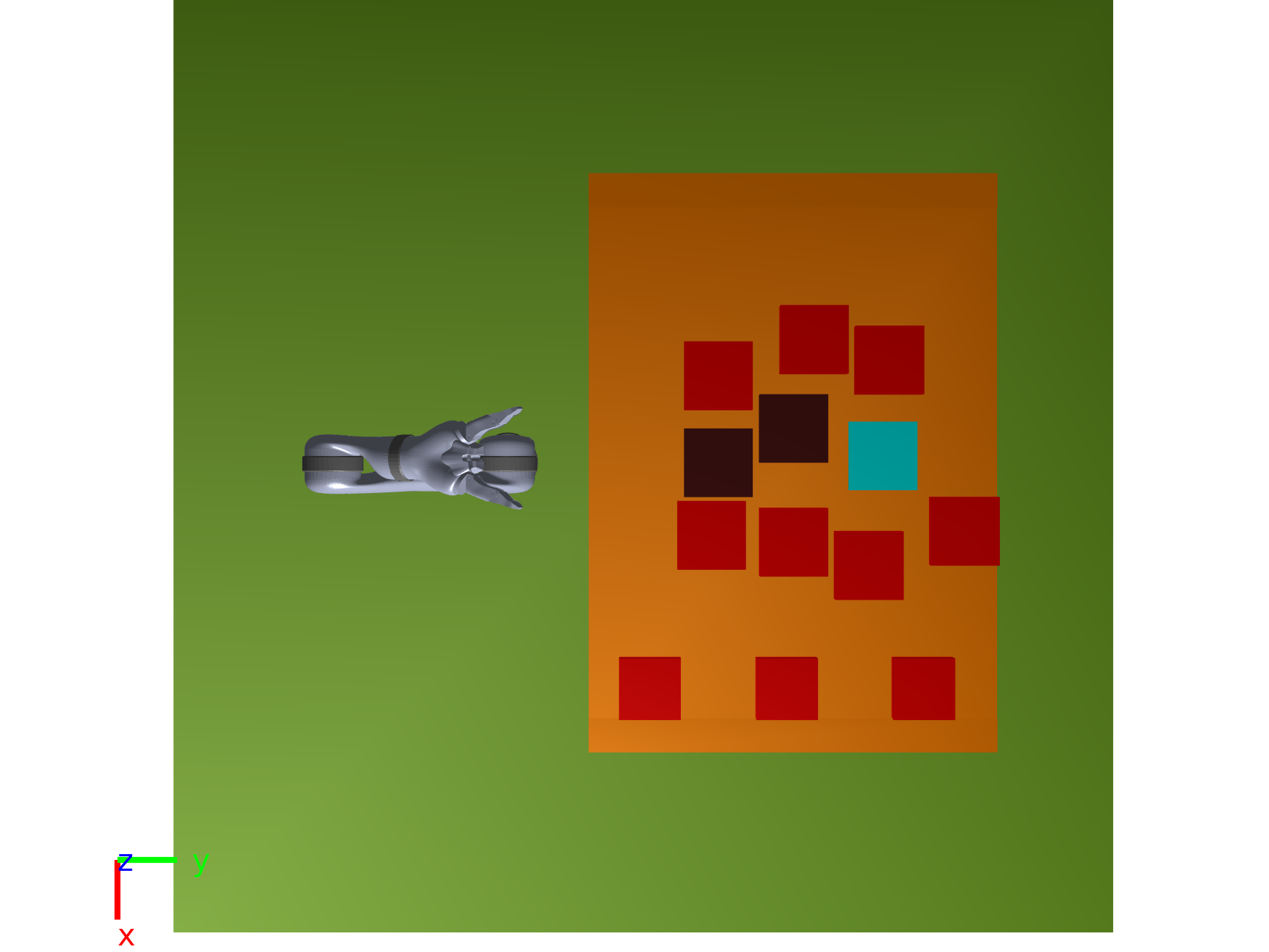}
\label{fig:manipulator_sim1}}
\hspace{0.05cm}
\subfloat[$S6$: Greedy search]{
\includegraphics[angle=90, trim=110 15 20 9,clip,scale=0.37]
{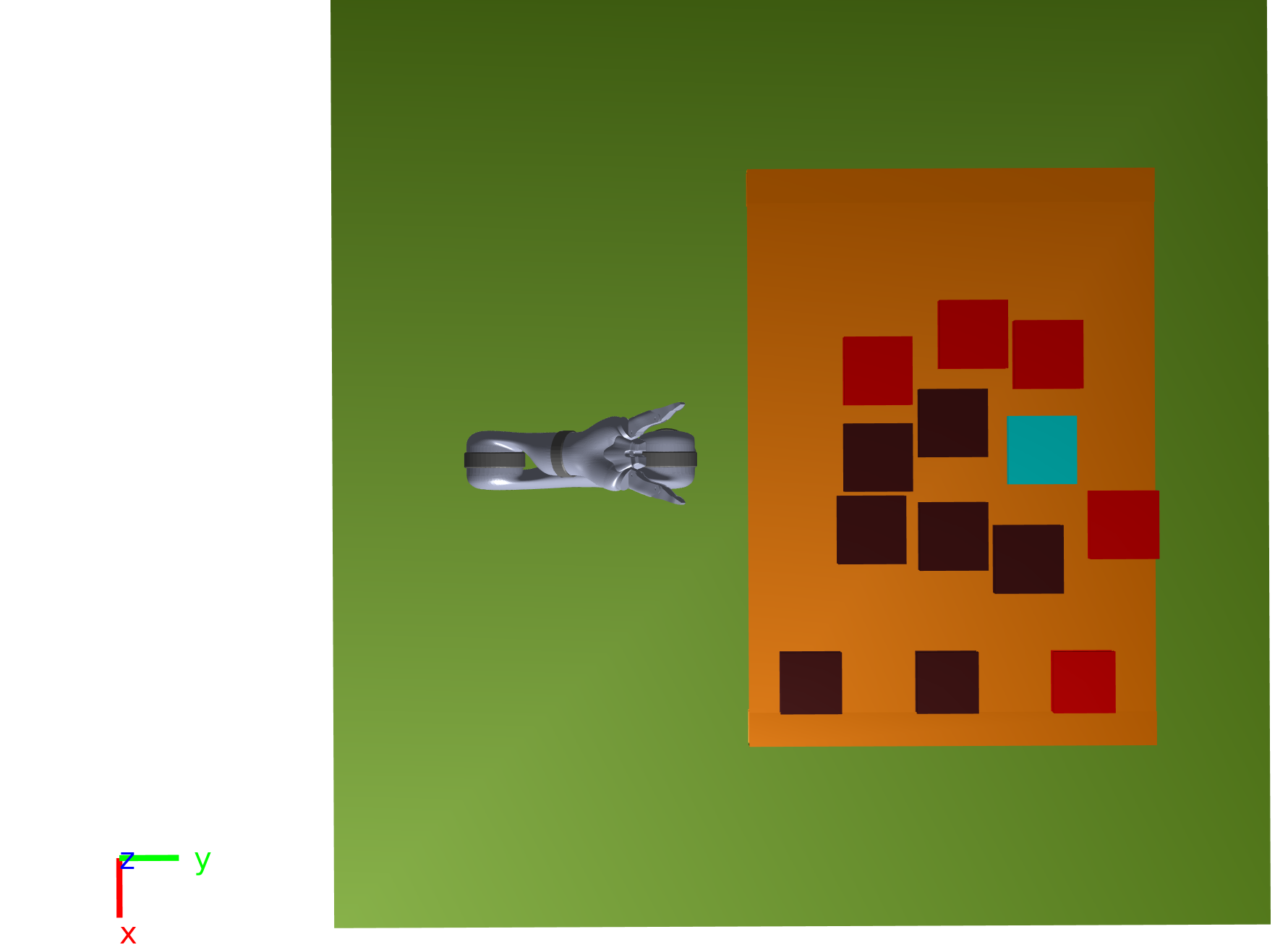}
\label{fig:manipulator_sim2}}
\caption{
Obstacle sets selected in the cluttered manipulation benchmark $S_6$. 
Selected objects are shown in \textit{brown}.
(a) Exact search and $\textsf{GFS-1}+\textsf{Refine}$ select
two objects.
(b) Greedy search selects seven objects.}
\label{fig:simulations2}
\end{figure}

We consider the following six benchmarks:
\begin{itemize}
\item
$S_1$--$S_4$:
a translating and rotating rectangular robot moves from left to right through environments containing removable triangular obstacles;
\item $S_5$:
a translating and rotating L-shaped robot moves through a larger environment containing removable triangular obstacles; and
\item $S_6$: 
a 4-DOF manipulator must reach and grasp a target object in clutter;
the removable constraints correspond to objects obstructing the arm motion.
\end{itemize}
Representative planar scenarios and their computed paths are shown in Figure~\ref{fig:simulations}.
Furthermore, Figure~\ref{fig:simulations2} compares the obstacle sets selected in $S_6$, using different approaches (ours, exact solution, and greedy).
It can be observed that exact search and $\textsf{GFS-1}+\textsf{Refine}$ both select two objects, whereas greedy search selects seven objects.
All six evaluated benchmarks satisfy the candidate-coverage condition addressed by \textsf{GFS-1}.
As a consequence, \textsf{GFS-2} is not invoked in these experiments. 
The discussed pipeline is therefore $\textsf{GFS-1}+\textsf{Refine}$. 
The role of \textsf{GFS-2} is to address the complementary case in which no candidate generated by \textsf{GFS-1} contains a complete optimal removal set, as analyzed in Section~\ref{subsec:gfs2}.

\noindent\textit{Graph construction.}
For each scenario, the robot configuration space is discretized over its degrees of freedom.
Each sampled configuration $q$ is assigned an obstacle label
\[
I(q)=\{o_i\in\mathcal O\mid q\text{ is in collision with }o_i\}.
\]
Configurations having the same label are grouped into connected components using grid adjacency. 
Each connected component becomes a vertex of the \textsf{MCR} graph, and an edge is introduced between components containing adjacent grid configurations.
The free-space components containing the start and goal configurations are represented by $s$ and $g$, respectively. 
The resulting graph is constructed once per scenario and shared by all evaluated
search methods.
Chosen resolutions are 
$\delta_x = 0.5$ units, 
$\delta_y = 0.5$ units, and
$\delta_\theta = 5^\circ$ for $S_1$--$S_5$, and 
$(\delta_1, \ldots, \delta_4) = (10^\circ, \ldots, 10^\circ)$ for $S_6$.

\begin{table*}[t]
\centering
\footnotesize
\setlength{\tabcolsep}{3.4pt}
\begin{tabular}{c cc c c c c cccc}
Scenario & $|V|$ & $|E|$ & $t_{\mathrm{E}}$ (s) & $t_{\mathrm{P}}$ (s) & Speed-up ($t_{\mathrm{E}}/t_{\mathrm{P}}$) & $t_{\mathrm{G}}$ (s) & $|S_{\mathrm{P}}|$ & $|S_{\mathrm{P}}|$ & $\rho$ & $|S_{\mathrm{G}}|$\\
\hline
\hline
$S_1$ & 2051 & 45446 & 478.62 &
52.34 (51.43+0.91) & 9.14 & 0.21 & 5 & 5 & 1.00 & 5\\
$S_2$ & 2051 & 45568 & 211.82 &
61.51 (60.53+0.98) & 3.44 & 0.58 & 5 & 5 & 1.00 & 27\\
$S_3$ & 2055 & 32479 & 140.69 &
37.08 (36.32+0.76) & 3.79 & 0.23 & 8 & 9 & 1.13 & 41\\
$S_4$ & 1488 & 35798 & 96.96 &
13.30 (12.02+1.28) & 7.29 & 0.67 & 9 & 11 & 1.22 & 23\\
$S_5$ & 7156 & 67535 & 1128.92 &
271.84 (269.38+2.46) & 4.15 & 0.33 & 16 & 21 & 1.31 & 30\\
$S_6$ & 794 & 314821 & 903.55 &
26.80 (26.71+0.09) & 33.71 & 0.55 & 2 & 2 & 1.00 & 7\\
\end{tabular}
\caption{
Graph-search results for the six benchmarks, averaged over 10 repeated runs. 
$E$, $P$, and $G$ denote exact search, $\textsf{GFS-1}+\textsf{Refine}$, and greedy search, respectively. 
Graph-construction time is excluded because all methods receive the same precomputed \textsf{MCR} graph.
Our method time is reported as $total\;(\textsf{GFS-1}+\textsf{Refine})$, and
$\rho = |S_{\mathrm{P}}|/|S^\star|$.
Run-to-run variation was negligible at the reported precision and is not shown.
}
\label{tab:statistics}
\end{table*}

\noindent\textit{Search performance.}
Table~\ref{tab:statistics} summarizes the results. 
Our approach returns an optimal solution in three of the six benchmarks, namely $S_1$, $S_2$, and $S_6$. 
In the remaining scenarios, the observed solution-cardinality ratios are $1.13$, $1.22$, and $1.31$, respectively. 
Thus, the worst observed ratio is $1.31$,
with a mean ratio of $1.11$ across the six benchmarks.

Compared with exact search, our approach is between $3.44$ and $33.71$ times faster, with a median speed-up of $5.72$. 
Most of its computation time is spent in \textsf{GFS-1}, since candidate refinement requires between $0.09$ and $2.46$~s in the reported scenarios. 
The largest speed-up occurs in the manipulation benchmark $S_6$, where the proposed method
matches the optimal two-object solution while reducing search time from $903.55$~s to $26.80$~s.

Greedy search is consistently faster, requiring less than $0.7$~s in every scenario, but its solution quality degrades substantially. 
In $S_2$ and $S_3$, greedy search selects more than five times the optimal number of obstacles. 
Except in $S_1$, our method selects fewer obstacles than greedy search.
The reduction ranges from $30.0\%$ in $S_5$ to $81.5\%$ in $S_2$. 
These results demonstrate the intended trade-off, that is, $\textsf{GFS-1}+\textsf{Refine}$ is slower than aggressive greedy search, but returns substantially smaller removal sets, while remaining considerably faster than exact search.

\noindent\textit{Impact on downstream manipulation planning.}
We examined how obstacle-set cardinality affects the planning effort required in $S_6$.
This experiment measures computational planning cost rather than physical execution time. 
For every selected object, we 
(i) compute a candidate \textit{grasp} configuration using inverse kinematics, and
(ii) generate a collision-free arm motion to the corresponding grasp pose. 
The reported measurements exclude
physical grasp execution and object relocation.

For the seven objects selected by greedy search, inverse kinematics and motion planning require a total of $84.17$~s.
Together with the $0.55$~s \textsf{MCR} search time, this gives a combined planning time of $84.72$~s. 
For the two objects selected by $\textsf{GFS-1}+\textsf{Refine}$, downstream
inverse kinematics and motion planning require $3.79$~s.
Including the $26.80$~s \textsf{MCR} search time gives a combined
planning time of $30.59$~s, approximately $2.8$ times lower than for greedy search.

This result illustrates an important system-level trade-off, that is, although greedy \textsf{MCR} search is faster in isolation, the
larger removal set can increase the planning effort required by
subsequent manipulation actions.
The selection of a smaller obstacle set
can therefore reduce the overall planning time for completing the task.

\section{Conclusions}
\label{sec:conclusion}
In this paper, we introduce $\textsf{GFS-1}+\textsf{Refine}$, a graph-based heuristic pipeline for computing feasible, low-cardinality solutions to the Minimum Constraint Removal problem.
We also derive \textsf{GFS-2}, an extended feasible-set generator for instances in which the candidates produced by \textsf{GFS-1} do not contain a complete optimal removal set.
The evaluated $\textsf{GFS-1}+\textsf{Refine}$ pipeline matched the exact solution in three of six benchmarks and exhibited a worst observed solution-cardinality ratio of $1.31$.
Compared with exact search, it achieves speed-ups between $3.44$ and $33.71$, with a median speed-up of $5.72$.

In the cluttered manipulation benchmark, our pipeline selects two obstructing objects, compared with seven selected by greedy search. 
This reduces the combined \textsf{MCR}, inverse kinematics, and motion planning time from $84.72$~s to $30.59$~s. 
The results indicate that accepting additional computation at the obstacle-selection stage can reduce the planning effort required by subsequent manipulation actions.

The approach has no worst-case approximation guarantee, and currently assigns equal cost to every object removal.
Moreover, it identifies which objects should be relocated but does not compute or execute their relocation motions. 
Future work 
(i) will include an empirical investigation of \textsf{GFS-2} on benchmark families specifically constructed to exhibit the candidate-coverage conditions that motivate the extension, and
(ii) will incorporate object-dependent manipulation costs, grasp and placement feasibility, and whole-body validation on a humanoid robot.



\bibliographystyle{IEEEtran}
\bibliography{References}

\end{document}